\documentclass[lettersize,journal]{IEEEtran}
\usepackage{amsmath,amsfonts}
\usepackage{algorithmic}
\usepackage{algorithm}
\usepackage{array}
\usepackage[caption=false,font=normalsize,labelfont=sf,textfont=sf]{subfig}
\usepackage{textcomp}
\usepackage{stfloats}
\usepackage{url}
\usepackage{verbatim}
\usepackage{graphicx}
\usepackage{cite}
\usepackage{multirow}
\usepackage{color}

\begin{document}

\title{Improving Nighttime Driving-Scene  Segmentation via Dual Image-adaptive Learnable Filters}

\author{Wenyu Liu, Wentong Li, Jianke Zhu,~\IEEEmembership{Senior Member,~IEEE,} Miaomiao Cui, Xuansong Xie, Lei Zhang,~\IEEEmembership{Fellow,~IEEE}
\thanks{Wenyu Liu, Wentong Li, and Jianke Zhu are with the College of Computer Science and Technology, Zhejiang University, Hangzhou, China. Jianke Zhu is corresponding author. e-mail: ({liuwenyu.lwy@zju.edu.cn, liwentong@zju.edu.cn, jkzhu}@zju.edu.cn)}
\thanks{Miaomiao Cui and Xuansong Xie are with Damo Academy, Alibaba Group, Hangzhou, China. e-mail: (miaomiao.cmm@alibaba-inc.com, xingtong.xxs@taobao.com)}
\thanks{Lei Zhang is with The HongKong Polytechnic University, HongKong, China. e-mail: (cslzhang@comp.polyu.edu.hk)}
}



\maketitle

\begin{abstract}
Semantic segmentation on driving-scene images is vital for autonomous driving. Although  encouraging performance has been achieved on daytime images, the performance on nighttime images are less satisfactory due to the insufficient exposure and the lack of labeled data. To address these issues, we present an add-on module called dual image-adaptive learnable filters (DIAL-Filters) to improve the semantic segmentation in nighttime driving conditions, aiming at exploiting the intrinsic features of driving-scene images under different illuminations. DIAL-Filters consist of two parts, including an image-adaptive processing module (IAPM) and a learnable guided filter (LGF). With DIAL-Filters, we design both unsupervised and supervised frameworks for nighttime driving-scene segmentation, which can be trained in an end-to-end manner. Specifically, the IAPM module consists of a small convolutional neural network with a set of differentiable image filters, where each image can be adaptively enhanced for better segmentation with respect to the different illuminations. The LGF is employed to enhance the output of segmentation network to get the final segmentation result. The DIAL-Filters are light-weight and efficient and they can be readily applied for both daytime and nighttime images. Our experiments show that DAIL-Filters can significantly improve the supervised segmentation performance on ACDC\_Night and NightCity datasets, while it demonstrates the state-of-the-art performance on unsupervised nighttime semantic segmentation on Dark Zurich and Nighttime Driving testbeds. Codes and models are available at \url{https://github.com/wenyyu/IA-Seg}.


\end{abstract}


\section{Introduction}

\IEEEPARstart{S}{emantic} segmentation aims to divide an image into several regions with the same object category. As a fundamental task in computer vision, semantic segmentation is widely used in autonomous driving~\cite{yang2018real}, indoor navigation~\cite{teso2020semantic, thomas2021self} and virtual reality~\cite{Li2016Combining}. By taking advantage of the powerful feature presentation using convolutional neural networks, deep learning-based semantic segmentation methods~\cite{he2016deep, lin2017refinenet, zhao2017pspnet, weng2021stage, ji2020encoder, sun2021gaussian}
have achieved encouraging results on the conventional daytime datasets~\cite{geiger2012we, Cordts2016Cityscapes}. 
However, these methods generalize poorly to the case of adverse nighttime lighting, which is critical for real-world applications such as autonomous driving. In this work, we focus our attention on semantic segmentation tasks in nighttime driving scenarios.

\begin{figure}[t]
\centering
\includegraphics[width=0.48\textwidth]{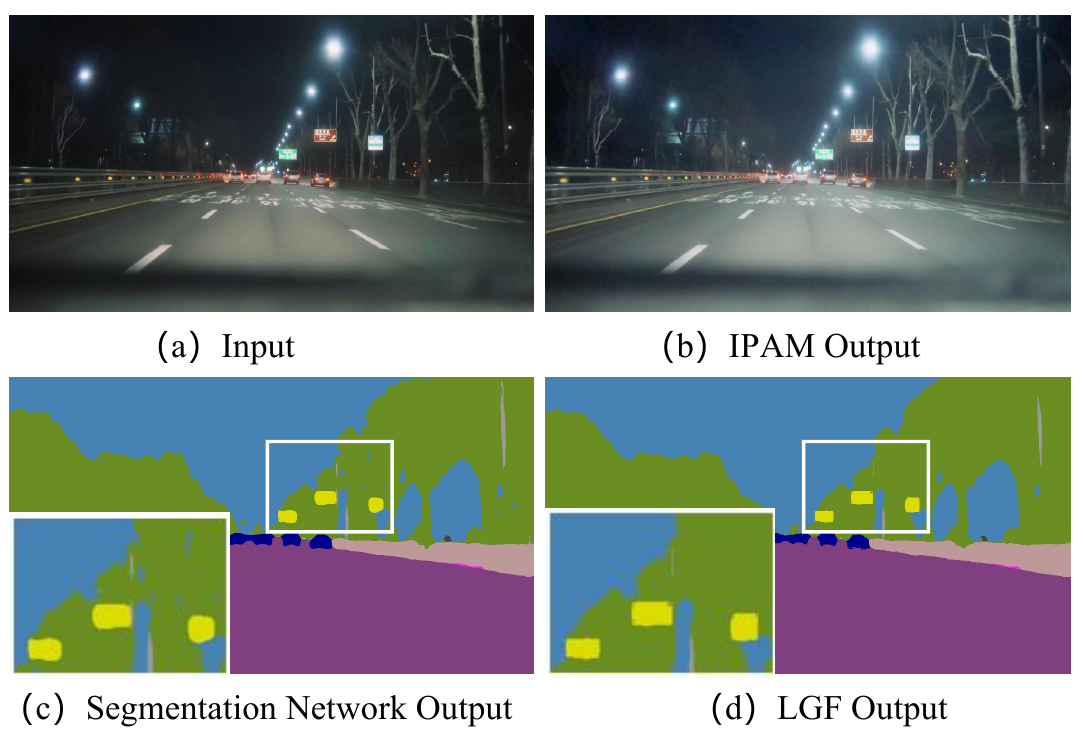} 
\caption{The visual results of different modules in our method. The IAPM module outputs clearer images with better brightness. The LGF module improves the segmentation performance on object boundaries.}
\label{fig:intro}
\end{figure}

There are two main challenges for nighttime driving-scene segmentation. One is the difficulty in obtaining the large-scale labeled nighttime datasets for the poor visual perception. To this end, several nighttime datasets have been developed recently~\cite{tan2021night, sakaridis2021acdc}. NightCity~\cite{tan2021night} contains 2,998 labeled nighttime driving-scene images and ACDC\_Night~\cite{sakaridis2021acdc} has 400 images, which can be used for supervised training. The other challenge is the exposure imbalance and motion blur in nighttime images, which is hard to deal with by the existing daytime segmentation methods. To tackle these challenges, some domain adaptation methods have been proposed to transfer the semantic segmentation models from daytime to nighttime without using labels in the nighttime domain. Domain adaptation network (DANNet)~\cite{wu2021dannet} employs adversarial learning for nighttime semantic segmentation, which adds an image relighting subnetwork before the segmentation network. This increases a large number of training parameters, which is not conducive to deployments. In~\cite{dai2018dark,sakaridis2019guided}, the twilight domain is treated as a bridge to achieve the domain adaptation from daytime to nighttime. Moreover, some methods~\cite{sakaridis2019guided,sun2019see,nag2019s,sakaridis2020map} take an image transfer model as a pre-processing stage to stylize nighttime or daytime images so as to construct synthetic datasets. By involving the complicated image transfer networks between day and night, these methods are usually computational intensive. In particular, it is difficult for the image transfer networks to achieve the ideal transformation when the inter-domain gap is large. 


The nighttime images captured in the driving scenes often contain both over-exposure and under-exposure parts, which seriously degrade the visual appearances and structures. Figure~\ref{fig:intro}(a) shows an example nighttime image with both over-exposure (street lights and car headlights) and under-exposure (background and trees) regions. Such uneven brightness deteriorates the image content and texture, making it difficult to accurately segment the object's boundary. In digital imaging systems, retouching experts improve the image quality by tuning the hyperparameters of an image enhancement module, including white balance adjustment, gamma correction, exposure compensation, detail enhancement, tone mapping, and etc. To avoid manually tuning these parameters, ``white-box" image-adaptive enhancement framework~\cite{hu2018exposure, yu2018deepexposure,zeng2020learning} is employed to improve the image quality.



To address the above issues, we propose an ingenious driving-scene semantic segmentation method to improve the performance via dual image-adaptive learnable filters (DIAL-Filters), including an image-adaptive processing module (IAPM) and a learnable guided filter (LGF) module.
Specifically, we present a set of fully differentiable image filters (DIF) in IAPM module, whose hyperparameters are adaptively predicted by a small CNN-based parameter predictor (CNN-PP)  according to the brightness, contrast and exposure information of the input image. Moreover, the LGF is suggested to enhance the output of segmentation network. A joint optimization scheme is introduced to learn the DIF, CNN-PP, segmentation network and LGF in an end-to-end manner. Additionally, we make use of both daytime and nighttime images to train the proposed network. By taking advantages of the CNN-PP network, our method is able to adaptively deal with  images of different lighting. Figure~\ref{fig:intro} shows an example segmentation process of our proposed approach.

Part of the above mentioned image-adaptive filtering techniques has been used in the detection task in our previous conference paper~\cite{liu2022image}. Comparing to~\cite{liu2022image}, we make the following new contributions in this work: 1) we extend the image-adaptive filtering methods to the nighttime segmentation task and achieve state-of-the-art results; 2) a learnable guided filter is proposed to improve the segmentation performance on object edge regions; 3) we develop both supervised and unsupervised segmentation frameworks.





The main contributions of this paper are summarized in threefold:
\begin{itemize}
\item We propose a novel lightweight add-on module, called DIAL-Filters, which can be easily added to the existing models. It is able to significantly improve the segmentation performance on nighttime images by double enhancement before and after the segmentation network.
\item We train our image-adaptive segmentation model in an end-to-end manner, which ensures that CNN-PP can learn an appropriate DIF to enhance the image for segmentation and learn a LGF to preserve edges and details.
\item The supervised experiments show that our proposed method can significantly improve segmentation performance on ACDC\_Night and NightCity datasets. The unsupervised experiments on Dark Zurich and Nighttime Driving testbeds show that our method achieves state-of-the-art performance for unsupervised nighttime semantic segmentation.  
\end{itemize}

\section{Related Work}
\subsection{Semantic Segmentation}
Image semantic segmentation is essential to many visual understanding systems, whose performance on benchmark datasets has been greatly improved  due to the development of Convolutional Neural Networks (CNNs). FCN~\cite{Long2015Fully} was considered as a milestone, which demonstrates the capability of training a deep network for semantic segmentation in an end-to-end manner on variable-size images. Multi-level-based methods~\cite{lin2017refinenet, zhao2017pspnet} employed the  multi-scale analysis to extract the global context while preserving the low-level details. Moreover, the convolution layer was used to generate the final per-pixel predictions. DeepLab and its variants~\cite{chen2017deeplab, chen2017rethinking, chen2018encoder} introduced Atrous Convolution and Atrous Spatial Pyramid Pooling to the segmentation network. 


All the above methods focus on segmentation in  daytime conditions. In this paper, we pay attention to night-time scenes. To investigate the effectiveness of our proposed DIAL-Filters on nighttime driving-scene segmentation, we select three popular and widely used segmentation networks as baselines, including RefineNet~\cite{lin2017refinenet}, PSPNet~\cite{zhao2017pspnet} and DeepLabV2~\cite{chen2017deeplab}.

\subsection{Image Adaptation}
Image adaptation is widely used in both low-level and high-level tasks. For image enhancement,  some traditional methods~\cite{polesel2000image, yu2004fast, wang2021adaptive} adaptively calculate the parameters of image transformation according to the corresponding image features. Wang~\textit{et al.}~\cite{wang2021adaptive} proposed an brightness adjustment function that adaptively tunes the enhancement parameters based on the illumination distribution characteristics of an input image. Methods in ~\cite{hu2018exposure,yu2018deepexposure,zeng2020learning} employed a small CNN to flexibly learn the hyperparameters of image transformation. Yu \textit{et al.}~\cite{yu2018deepexposure} utilized a small CNN to learn image-adaptive  exposures with deep reinforcement learning and adversarial learning. Hu \textit{et al.}~\cite{hu2018exposure} proposed a post-processing framework with a set of differentiable filters, where deep reinforcement learning (DRL) is used to generate the image operation and filter parameters according to the quality of the retouched image. For high-level detection task, Zhang \textit{et al.}~\cite{zhang2015image} presented an improved canny edge detection method that use mean values of gradient of entire image to adaptively select dual-threshold. IA-YOLO~\cite{liu2022image} proposed a light CNN to adaptively predict the filter's parameters for better detection performance.
Inspired by these methods, in this work we adopt image adaptation for segmentation in nighttime driving scenarios.

\subsection{Domain Adaptation}
 Domain adaptive methods~\cite{tian2021partial, zhang2018unsupervised, lin2020cross, bo2022all, tsai2018learning} have achieved encouraging performance in many tasks, such as classification, object detection, pedestrian identification and segmentation. The semantic segmentation methods with domain adaptation can be roughly divided into three categories, including adversarial learning~\cite{tsai2018learning, pan2020unsupervised, vu2019advent}, self-training~\cite{xie2022sepico, zhang2021prototypical, li2019bidirectional} and curriculum learning~\cite{zhang2017curriculum, lian2019constructing}. 
 
The adversarial learning-based methods reduced the distribution shift of two domains by means of adversarial training. AdaptSegNet~ \cite{tsai2018learning}  proposed a multi-level adversarial network which perform the output space domain adaption effectively at different feature levels. Some methods~\cite{pan2020unsupervised, vu2019advent} also addressed the unsupervised domain adaptation segmentation by means of the entropy scheme based on pixel-level prediction. The self-training methods utilized the unlabeled target data by training with pseudo labels generated from the pre-trained model in source domain. To name a few, Zhang \textit{et al.}~\cite{zhang2021prototypical} investigated the feature distances of prototypes to fine-tune the pseudo labels and distilled the pre-trained knowledge to a self-supervised model. Xie \textit{et al.}~\cite{xie2022sepico} proposed a one-stage end-to-end adaptive network for domain alignment using semantic-guided 
pixel contrast learning. The curriculum learning-based approaches~\cite{zhang2017curriculum, lian2019constructing} exploited the curriculum learning manner to learn the easy attributes of the target domain, and then employed it to regularize the final segmentation network.

However, most of these domain adaptation methods pay attention to the adaption of synthetic-to-real (i.e., GTA5~\cite{richter2016playing} to Cityscapes) or cross-city (i.e., Cityscapes to Cross-City~\cite{chen2017no}), which are all day-to-day adaptations. Therefore, these methods are often unable to properly deal with the significant adaptation gap between daytime and nighttime images, which cannot achieve satisfactory performance in nighttime segmentation~\cite{wu2021dannet}. In this paper, we focus on adaptation between daytime and nighttime domains.

\subsection{Nighttime Driving-scene Semantic Segmentation}
While most of the existing works focus on the ``normal'' scenarios with well-illuminated scenes, there are some studies to address the challenging scenarios such as nighttime scenes. Some researchers employed domain adaption-based methods~\cite{dai2018dark, sakaridis2019guided, sakaridis2020map} to transfer the model trained in normal scenes to the target domain. In~\cite{dai2018dark}, a progressive adaptation approach was proposed to transform from daytime to nighttime via the bridge of twilight time. 
Sakaridis \textit{et al.}~\cite{sakaridis2019guided,sakaridis2020map} presented a guided curriculum adaptation method based on DMAda~\cite{dai2018dark}, which gradually adapt segmentation models from day to night using both the annotated synthetic images and unlabeled real images.
However, the additional segmentation models for different domains in these gradual adaptation methods increase the computation cost significantly.
Some studies~\cite{romera2019bridging,sun2019see,nag2019s} trained the additional style transfer networks, e.g., CycleGAN~\cite{zhu2017unpaired}, to perform the day-to-night or night-to-day image transfer before training the semantic segmentation models. The disadvantage of these methods is that the performance of the subsequent segmentation network lies in highly dependent on the previous style transfer model.


Recently, Wu \textit{et al.}~\cite{wu2021dannet,wu2021one} proposed an unsupervised one-stage adaptation method, where an image relighting network is placed at the head of the segmentation network. Adversarial learning was employed to achieve domain alignment between the labeled daytime data and unlabeled nighttime data. Unfortunately, the additional RelightNet incurs a large number of parameters and computation.

In contrast to the above methods, we suggest an image-adaptive segmentation approach to nighttime segmentation by embedding the proposed DIAL-Filters into a segmentation network. Our method can also be trained on unsupervised domain adaption with adversarial loss, which demonstrates significant advantages in both performance and efficiency.

\begin{figure*}[t]
\centering
\includegraphics[width=0.9\textwidth]{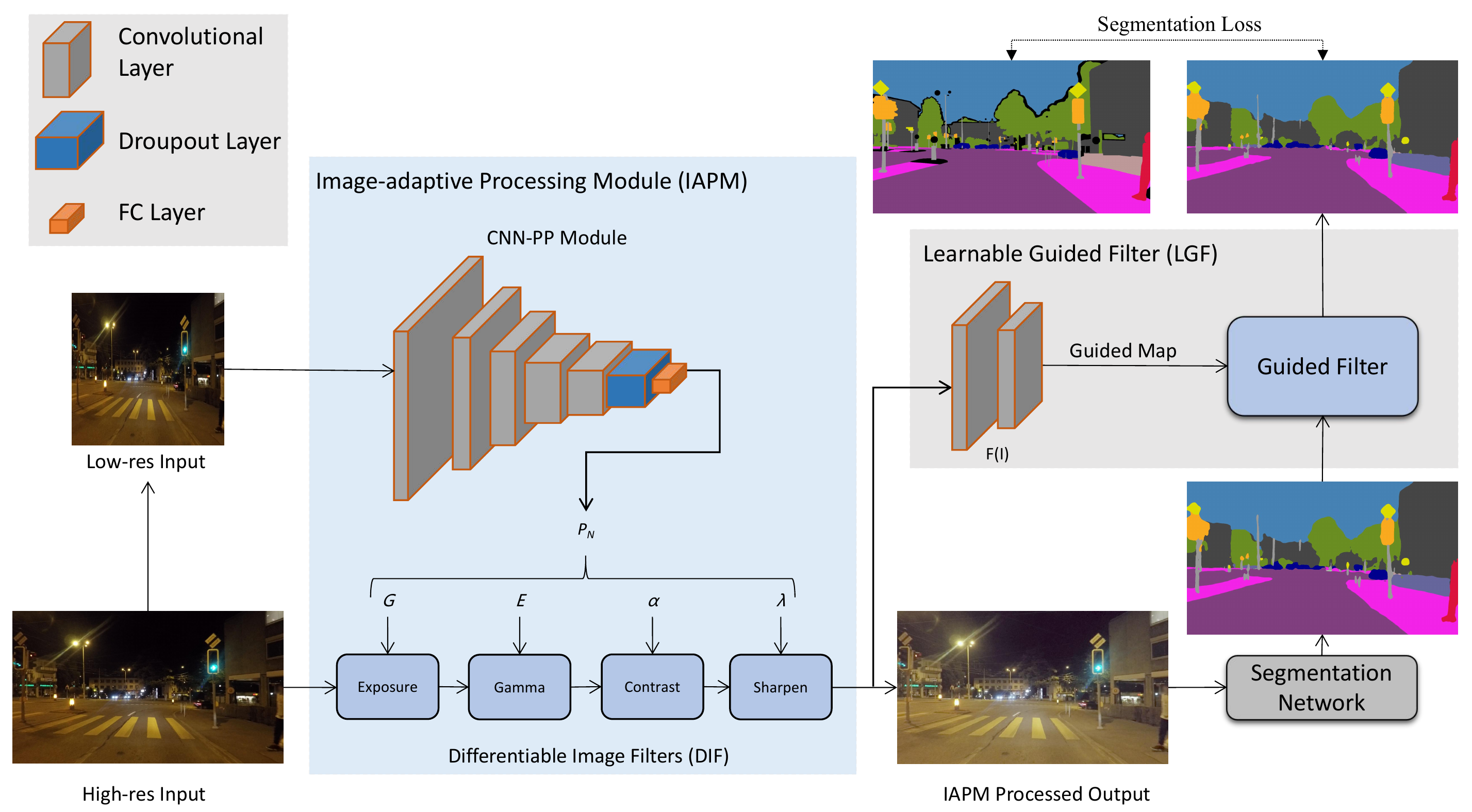} 
\caption{The end-to-end training pipeline of our proposed supervised segmentation framework. Our method learns a segmentation network with a small CNN-based parameter predictor (CNN-PP), which employs the downsampled input image to predict the hyperparamters of filters in the DIF module. The input high-resolution images are processed by DIF to aid segmentation network for better segmentation performance.}
\label{fig:BIF-pipeline}
\end{figure*}

\section{Dual Image-adaptive Learnable Filters }
Driving-scene images captured in nighttime have poor visibility due to the weak illuminations, which lead to the difficulties in semantic segmentation. Since each image may contain both overexposed and underexposed regions, the key of alleviating the difficulty in nighttime segmentation is to deal with exposure difference. Therefore, we suggest a set of  dual image-adaptive learnable filters (DIAL-Filters) to enhance the results before and after the segmentation network. As illustrated in Figure~\ref{fig:BIF-pipeline}, the whole pipeline consists of an image-adaptive processing module (IAPM), a segmentation network and a learnable guided filter (LGF). The IAPM module includes a CNN-based parameters predictor (CNN-PP) and a set of differentiable image filters (DIF).


\subsection{Image-adaptive Processing Module} 
\subsubsection{Differentiable Image Filters}
As in \cite{hu2018exposure}, the design of image filters should conform to the principle of differentiability and resolution-Independence. For the gradient-based optimization of CNN-PP, the filters should be differentiable in order to allow the network training by backpropagation. Since CNN may consume intensive computational resources to process high resolution images (e.g., $4000 \times 3000$), we learn the filter parameters from the downsampled low-resolution image of size $256\times 256$. Moreover, the same filter is applied to the image of original resolution so that these filters are independent of image resolution.


Our proposed DIF consists of several differentiable filters with adjustable hyperparameters, including \emph{Exposure}, \emph{Gamma}, \emph{Contrast} and \emph{Sharpen}. As in~\cite{hu2018exposure}, the standard color operators, such as \emph{Gamma}, \emph{Exposure} and  \emph{Contrast}, can be expressed as pixel-wise filters. 

\emph{Pixel-wise Filters.} The pixel-wise filters map an input pixel value $P_{i} = (r_{i}, g_{i}, b_{i})$ into an output pixel value $P_{o} = (r_{o}, g_{o}, b_{o})$, in which $(r, g, b)$ represent the values of the three color channels of red, green and blue, respectively. The mapping functions of the three pixel-wise filters are listed in Table \ref{tab:map-fuction}, where the second column lists the parameters to be optimized in our approach. \emph{Exposure} and \emph{Gamma} are simple multiplication and power transformations. Obviously, these mapping functions are differentiable with respect to both the input image and their parameters. 

\begin{table}[t]
\caption{The mapping functions of pixel-wise filters\label{tab:map-fuction}}
\centering
    \small
    \begingroup
    \setlength{\tabcolsep}{1.0pt} 
    \renewcommand{\arraystretch}{1} 
{
\begin{tabular}{lll}
    \hline
    Filter & Parameters & Mapping Function  \\
    \hline
    Exposure & $E$: Exposure value & $P_{o} = 2^{E} \cdot P_{i} $ \\
    Gamma & $G$: Gamma value & $P_{o} = P_{i}^{G}$  \\
    Contrast & $\alpha $: Contrast value & $P_{o} = \alpha \cdot En(P_{i})+(1-\alpha) \cdot P_{i}$ \\
    \hline
\end{tabular}
}
    \endgroup

\end{table}

The differentiable contrast filters are designed with an input parameter to set the linear interpolation between the original image and the fully enhanced image. As shown in Table~\ref{tab:map-fuction}, the definition of $En(P_{i})$ in the contrast filter mapping function is as follows:
\begin{equation}
En(P_{i})=P_{i}\times\frac{EnLum(P_{i})}{Lum(P_{i})},\label{1}
\end{equation}
where $Lum(P_{i})$ is the luminance function based on the sensitivity of the human eye to the three primary colours, and $EnLum(P_{i})$ is the enhanced  luminance function. They are defined as follows:
\begin{equation}
Lum(P_{i}) = 0.27r_{i}+0.67g_{i}+0.06b_{i}, \label{2}
\end{equation}
\begin{equation}
EnLum(P_{i})=\frac{1}{2}(1-\cos(\pi \times (Lum(P_{i})))).\\\label{3}
\end{equation}

\emph{Sharpen Filter.} Image sharpening can highlight the image details. Like the unsharpen mask technique~\cite{polesel2000image}, the sharpening process can be described as follows:
\begin{equation}
F(x, \lambda) = I(x) + \lambda(I(x)- Gau(I(x))),\label{4}\\
\end{equation} 
where $I(x)$ is the input image. $Gau(I(x))$ denotes Gaussian filter, and $\lambda$ is a positive scaling factor. This sharpening operation is differentiable with respect to both $x$ and $\lambda$. Note that the sharpening degree can be tuned for better segmentation performance by optimizing $\lambda$.

\subsubsection{CNN-based Parameters Predictor (CNN-PP)}
In camera image signal processing (ISP) pipeline, some adjustable filters are usually employed for image enhancement, whose hyperparameters are manually tuned by experienced engineers through visual inspection~\cite{mosleh2020hardware}. Such a tuning process is very awkward and expensive to find the suitable parameters for a broad range of scenes. To address this limitation, we employ a small CNN as a parameter predictor to estimate the hyperparameters, which is very efficient.

The purpose of CNN-PP is to predict the DIF's parameters by understanding the global content of the image, such as brightness, color and tone, as well as the degree of exposure. The downsampled image is sufficient to estimate such information, which can greatly save the computational cost. 
As in~\cite{zeng2020learning}, we apply a small CNN-PP to the low-resolution version of the input image to predict the hyperparameters of DIF. Given an input image of any resolution, we simply use bilinear interpolation to downsample it into $256 \times 256$ resolution. As shown in Figure~\ref{fig:BIF-pipeline}, the small CNN-PP network is composed of five convolutional blocks, and the final fully-connected layer outputs the hyperparameters for the DIF module. The CNN-PP model contains only 278K parameters when the total number of DIF’s hyperparameters is 4.


\subsection{Learnable Guided Filter }
Many recent approaches to high-level visual tasks cascade a guided filter behind their original architecture to improve the results~\cite{hu2017deep, wu2018fast}. Guided filter~\cite{he2012guided}  is a type of edge-preserving and gradient-preserving image operation, which makes use of the object boundary in the guidance image to detect the object saliency. It is able to suppress the saliency outside the objects, improving the down-streamed detection or segmentation performance.
\begin{center}
\begin{figure*}[t]
\centering
\includegraphics[width=0.9\textwidth]{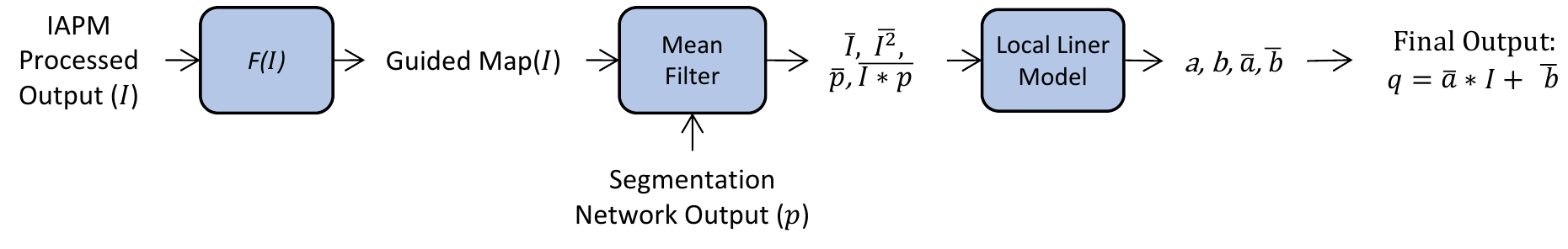} 
\caption{The pipeline of learnable guided filter (LGF). The LGF module takes the IAPM output $I$ and segmentation network result $p$ as inputs, and outputs the enhanced segmentation result. With $F(I)$, we can adaptively process each image for better segmentation performance with edge preservation.}
\label{fig:LGF-piepline}
\end{figure*}    
\end{center}

The original guided filter has a guided map $I$, an input image $p$, and the output image $q$. As in Eq.~(\ref{5}), it supposes that $q$ is a linear transformation of $I$ in a window $\omega_{k}$ centered at the pixel $k$.
\begin{equation}
q_{i} = a_{k}I_{i} + b_{k}, \forall i \in \omega_{k}\label{5}.\\
\end{equation} 
The ($a_{k}, b_{k}$) are some linear coefficients which are assumed to be
constants in $\omega_{k}$. The $\omega_{k}$ is a square window of a radius $r$. We can obtain the final solution of ($a_{k}, b_{k}$) as follows~\cite{he2012guided}
\begin{equation}
a_{k} =\frac{\frac{1}{|\omega|}\sum_{i \in \omega_{k}}I_{i}p_{i} - \mu_{k}\bar{p}_{k}}{\sigma ^ 2 _{k} + \epsilon},\label{6}
\end{equation}
\begin{equation}
b_{k} =\bar{p}_{k} - a_{k}\mu_{k},\label{7}
\end{equation}
where $\mu_{k}$ and $\sigma ^ 2 _{k}$ are the mean and variance of $I$ in a window $\omega_{k}$. $|\omega|$ is the number of pixels in $\omega_{k}$, $\epsilon$ is a regularization parameter, and $\bar{p}_{k} = \frac{1}{|\omega|}\sum_{i \in \omega_{k}}p_{i}$ is the mean of $p$ in $\omega_{k}$.
When applying the linear transformation to each window $\omega_{k}$, as shown in Eq.~(\ref{8}), we can obtain the filtering output by averaging all the possible values of $q_{i}$:
\begin{equation}
q_{i} = \bar{a}_{i}I_{i} + \bar{b}_{i},\label{8}\\
\end{equation}
where $\bar{a}_{i} = \frac{1}{|\omega|} \sum_{k \in \omega_{i}}a_{k} $ and $\bar{b}_{i} = \frac{1}{|\omega|} \sum_{k \in \omega_{i}}b_{k}$ are the average coefficients of all windows overlapping $i$. To further enhance the segmentation results, we introduce a learnable guide filter (LGF) behind the segmentation network. Algorithm~\ref{alg1} is the pseudo code of our LGF module, in which $f_{mean}$ denotes a mean filter with a window radius $r$.
The abbreviations of correlation (corr), variance (var), and
covariance (cov) represent the original meaning of these
variables. The detailed derivation process can be found in~\cite{he2012guided}. Figure~\ref{fig:LGF-piepline} illustrates LGF's architecture. The input $p$ is the output of the segmentation network, which has 19 channels.  The guided map $I$ is the output of $F(I)$. $F(I)$ involves with two convolutional layers having 64 and 19 output channels, containing only 1,491 parameters. It ensures that both $I$ and $p$ have the same number of channels. The LGF module is trained along with other modules in an end-to-end manner, which ensures LGF to adaptively process each image for better segmentation performance with edge preservation. 

\begin{algorithm}[t]
  \caption{Learnable Guided Filter Procedure}
  \label{alg1}
  \begin{algorithmic}
\STATE \textbf{Input:} segmentation network output $p$, IAPM processed output $I$, radius $r$, regularization $\epsilon$ 

\STATE \textbf{Output:} filtering output $q$ \\

\STATE \textbf{1:} $I = F(I)$\\
\STATE \textbf{2:} $\bar{I} = mean_I = f_{mean}(I)$\\ ~~~$\bar{p} = mean_p = f_{mean}(p)$\\
~~~$\overline{I^{2}} = corr_I = f_{mean}(I \ast I)$\\ ~~~$\overline{I \ast p} = corr_{Ip} = f_{mean}(I \ast p)$ 
\STATE \textbf{3:} $var_I = corr_I - mean_I \ast mean_I$\\
~~~$cov_{Ip} = corr_{Ip} - mean_I \ast mean_p$\\
\STATE \textbf{4:} $a = cov_{Ip} /(var_I +\epsilon )$\\
~~~$b = mean_p - a \ast mean_I$\\
\STATE \textbf{5:} $\bar{a} = f_{mean}(a)$\\
~~~$\bar{b} = f_{mean}(b)$\\
 \STATE \textbf{6:} $q = \bar{a} \ast I + \bar{b}$
  \end{algorithmic}
\end{algorithm}

\section{Nighttime Semantic Segmentation }
The proposed DIAL-Filters are added to the segmentation network to form our nighttime segmentation method. As shown in Figure~\ref{fig:BIF-pipeline}, we plugin the IAPM and LGF into the head and end of the segmentation network, respectively. Most of existing methods adopt the unsupervised domain adaption methods to deal with nighttime segmentation. To make a more comprehensive comparison, we propose both supervised and unsupervised segmentation frameworks based on DIAL-Filters in this paper. 

\subsection{Supervised Segmentation with DIAL-Filters}
\subsubsection{Framework}
 As illustrated in Figure~\ref{fig:BIF-pipeline}, our supervised nighttime segmentation method consists of an IAPM module, a segmentation network and a LGF module. The IAPM module includes a CNN-based parameters predictor (CNN-PP) and a set of differentiable image filters (DIF). We firstly resize the input image into the size of $256\times 256$, and feed it into CNN-PP to predict DIF's parameters. Then, the image filtered by DIF is treated as the input for segmentation network. The preliminary segmentation image is filtered by LGF to obtain the final segmentation results. The whole pipeline is trained end-to-end with segmentation loss so that the CNN-PP is able to learn an appropriate DIF to enhance the image adaptively for better semantic segmentation.  
\subsubsection{Segmentation Network}
Following~\cite{wu2021dannet}, we select three popular semantic segmentation networks in our method, including DeepLabV2~\cite{chen2017deeplab}, RefineNet~\cite{lin2017refinenet} and PSPNet~\cite{zhao2017pspnet}. All these methods are with the common ResNet-101 backbone~\cite{he2016deep}.
\subsubsection{ Re-weighting and Segmentation Loss}
Since the numbers of pixels for different object categories in the driving-scene images are  uneven, it is difficult for the network to learn the features for the categories of small-size objects. This leads to  poor performance in predicting the pixels of small objects. Following~\cite{wu2021dannet}, we use a re-weighting scheme to improve the network's attention to small-size objects. The re-weighting equation is defined as follows:
\begin{equation}
w'_m = -\log(a_m), \label{9}\\
\end{equation}
where $a_m$ represents the proportion of pixels which are annotated as category $m$ in the labeled Cityscapes dataset. Obviously, the lower the value of $a_m$ is, the higher the weight $w'_m$ is assigned. Therefore, it facilitates the network to segment the categories of smaller-size objects.
For each category $m\in \mathbb{K}$, the weight $w'_m$ is normalized as follows: 
\begin{equation}
w_m = \frac{w'_m -\overline{w}}{\sigma(w)}\cdot e + 1.0,
\label{10}\\
\end{equation}
where $e$ is an adjustable hyperparameter, $\overline{w}$ and $\sigma(w)$ are the mean and standard deviation of $w'_m$, respectively. We set $e=0.05$ by default during training.

We utilize the popular weighted cross-entropy loss to account for segmentation:
\begin{equation}
L_{seg} = -\frac{1}{N |\mathbb{K}|}\sum _{m\in \mathbb{K}}\|w_m GT^{(m)} \cdot \log(P^{(m)}) \|_1, \label{11}\\
\end{equation}
where $P^{(m)}$ is the $m$-th channel of the segmentation result, $w_m$ is the weight set in Eq.~(\ref{10}). $N$ is the number of valid pixels in the corresponding segmentation labeled image, $|\mathbb{K}|$ is the number of labeled categories in the Cityscapes dataset, and $GT^{(m)}$ denotes the one-hot encoding of the ground truth of the $m$-th category.

\begin{figure*}[t]
\centering
\includegraphics[width=1.0\textwidth]{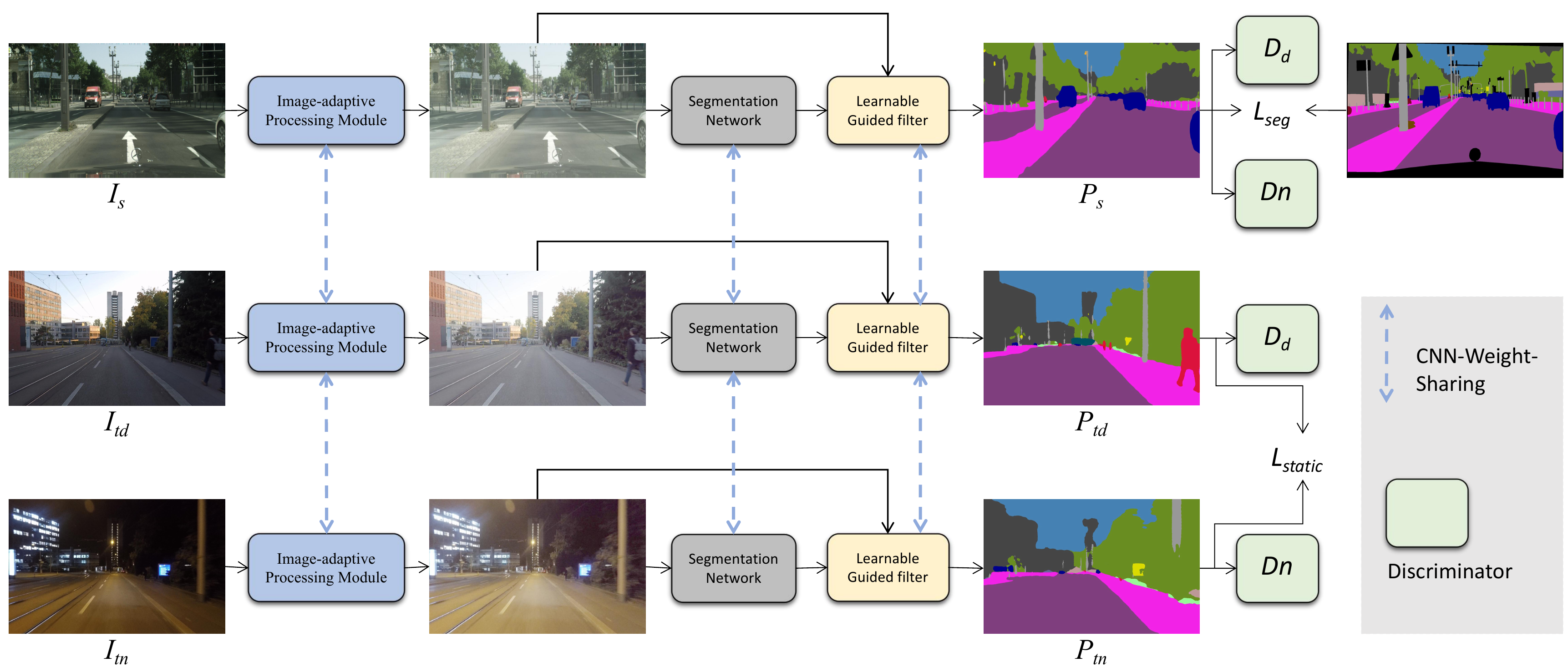} 
\vspace{-2mm}
\caption{The end-to-end training pipeline of our proposed unsupervised segmentation framework. Three images ($I_s$, $I_{td}$ and $I_{tn}$) from daytime Cityscapes (source domain $S$) , Dark Zurich-Daytime (daytime target domain $T_d$) and Dark Zurich-nighttime (nighttime target domain $T_n$) are input to the weight-sharing IAPM. Then, the enhanced outputs are fed into the weight-sharing segmentation network to obtain the preliminary segmentation results. Finally, the segmentation predictions are filtered by a weight-sharing LGF to get the final results. The corresponding IAPM outputs the guided map of LGF.
}
\label{fig:unsuperpipeline}
\end{figure*}

\subsection{Unsupervised Segmentation with DIAL-Filters}

\subsubsection{Framework}
The Dark Zurich~\cite{GCMA_UAE} is a relatively comprehensive nighttime dataset for real-world driving scenarios, which contains the corresponding images of the same driving scenes at daytime, twilight and nighttime. There are three domain images in our unsupervised method, including a source domain $S$ and two target domains $T_d$ and $T_n$, where $S$, $T_d$, and $T_n$ denote Cityscapes (daytime), Dark Zurich-D (daytime), and Dark Zurich-N (nighttime), respectively. As shown in Figure~\ref{fig:unsuperpipeline}, our unsupervised nighttime segmentation framework employs the similar architecture to~\cite{wu2021dannet}. The proposed unsupervised framework consists of three training circuits, which executes domain adaption from labeled source domains $S$  to two target domains $T_d$ and $T_n$ through the weight-sharing IAPM module, segmentation network and LGF module. It is worthy mentioning that only the images in Cityscapes have the semantic labels during training.
 
\subsubsection{Discriminators}
Following~\cite{tsai2018learning}, we design the discriminators to distinguish whether the segmentation results are from the source domain of the target domains by applying adversarial learning. Specifically, there are two discriminators with the same structures in our model. Each of them involves five convolutional blocks with the output channel of $\{64, 128, 256, 256, 1\}$. Each convolutional block includes a 4 × 4 convolution layer with a Leaky Relu. Except that the stride of the first two convolution layers is 2, the rest is 1. They are trained to distinguish whether the output is from $S$ or $T_d$ and from $S$ or $T_n$, respectively.

\subsubsection{Objective Functions} When training the proposed end-to-end unsupervised framework, we use the total loss $L_{total}$ for generator and the corresponding adversarial  loss for discriminator. The total loss $L_{total}$ consists of segmentation loss $L_{seg}$, static loss $L_{static}$ and adversarial loss $L_{adv}$.  

\emph{Segmentation Loss:} As in Eq.~(\ref{11}), we take the weighted cross-entropy loss as the segmentation loss. In particular, in our unsupervised framework, only the annotated source domain images are used to optimize this loss. We also set $std=0.05$ and $avg=1.0$ during the unsupervised training process.

\emph{Static Loss:} Considering the similarities between the daytime images in Dark Zurich-D and their corresponding nighttime images in Dark Zurich-N, we employ a static loss for the target domain $T\_n$ nighttime images as in~\cite{wu2021dannet}. This supports pseudo pixel-level supervision for the static object categories, e.g., road, sidewalk, wall, vegetation, terrain and sky. 

We first define $P_{td} \in \mathbb{R}^{H \times W \times C}$ as the target domain daytime segmentation result. $P_{tn} \in \mathbb{R}^{H \times W \times C} $ represents the corresponding nighttime segmentation prediction. When calculating the static loss, we only pay attention to the channels corresponding to the static categories. Thus, we can obtain  $P_{td}^\mathcal S \in \mathbb{R}^{H \times W \times C^ \mathcal S}$ and $P_{tn}^\mathcal S \in \mathbb{R}^{H \times W \times C^ \mathcal S}$, where $C^\mathcal S$ is the number of the categories of static objects. We then obtain the re-weighted daytime segmentation result $F_{td}$ as the pseudo label by Eq.~(\ref{10}). Finally, the static loss $L_{static}$ is defined as below:
\begin{equation}
L_{static} = -\frac{1}{N}\|(1 - P_{tn}^\mathcal S) \cdot \log(\tau)\|_1, \label{12}
\end{equation}
where $N$ is the number of valid pixels in the corresponding segmentation labeled map. $\tau$ denotes the likelihood map of the correct category, which is defined as follows:
\begin{equation}
\tau(c,i) = \max_{j}( o(c,j) \cdot P_{tn}^\mathcal S(c,i)).\label{13}
\end{equation}
The operation $o$ represents the one-hot encoding of the semantic pseudo ground truth $F_{td}$, and $j$ is each position of the $3 \times 3$ window centered at $i$.

\emph{Adversarial Loss:} Generative adversarial training is widely used to align two domains. In this case, we use two discriminators to distinguish whether the segmentation prediction is from the source domain or the target domain. We employ the least-squares loss function~\cite{mao2017least} in our adversarial training. The adversarial loss is defined by:
\begin{equation}
 L_{adv} = (D_{d}(P_{td})-s)^2 + (D_{n}(P_{tn})-s)^2,\label{14}
\end{equation}
where $s$ is the label for the source domain. Finally, we define the total loss $L_{total}$ of the generator (G) as follows:
\begin{equation}
\min\limits_G L_{total} = \alpha_1 L_{seg} + \alpha_2 L_{static} +\alpha_3 L_{adv},\label{15}
\end{equation}
where $\alpha_1, \alpha_2$ and $\alpha_3$ are set to 1, 1 and 0.01, respectively, during training.

The loss functions of two discriminators $D_s$ and $D_n$ are defined as follows:
\begin{equation}
\begin{aligned}
\min\limits_{D_{d}}L_{d}=  \frac{1}{2} (D_{d}(P_s)-s)^2+ 
 \frac{1}{2} (D_{d}(P_{td})-t)^2,   
\end{aligned}
\end{equation}
\vspace{-0.2cm}
\begin{equation}
\begin{aligned}
\min\limits_{D_{n}}L_{n} =  \frac{1}{2} (D_{n}(P_s)-s)^2 +  \frac{1}{2} (D_{n}(P_{tn})-t)^2,   
\end{aligned}
\end{equation}
where $t$ is the label for the target domains. 

\section{Experiments}
In this section, we first present the experimental testbeds and evaluation metrics. Then, we perform both unsupervised and supervised experiments to investigate the effectiveness of our method in nighttime driving-scene semantic segmentation. For the supervised experiments, we evaluate our approach on three datasets, including Cityscapes~\cite{Cordts2016Cityscapes},  NightCity~\cite{tan2021night} and ACDC~\cite{sakaridis2021acdc}, which have ground truth with pixel-level semantic annotations. For the unsupervised tests, we perform a domain adaption from Cityscapes (with labels) to Dark Zurich~\cite{GCMA_UAE}.

\begin{figure*}[t]
\centering
\includegraphics[width=0.95\textwidth]{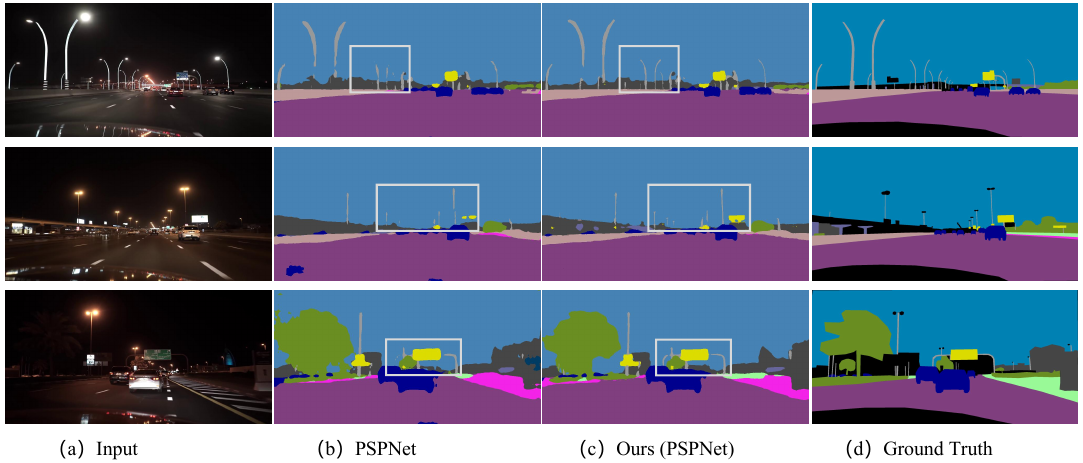} 
\caption{Visual segmentation results of our method and baseline model on NightCity\_test images. All the methods are trainned on Cityscapes and NightCity.}
\label{fig:nightcity_viusal}
\end{figure*}

\subsection{Datasets and Evaluation Metrics} For all experiments, we employ the mean of category-wise intersection-over-union (mIoU) as the evaluation metric. The following datasets are used for model training and performance evaluation.

\subsubsection{Cityscapes~\cite{Cordts2016Cityscapes}} 
Cityscapes is a semantic understanding dataset focused on daytime urban street scenes, which is widely used as a benchmark dataset for segmentation tasks. It includes 19 categories of pixel-level annotations, and consists of 2,975 training images, 500 validation images and 1,525 testing images with $2,048 \times 1,024$ resolution. In this work, we employ Cityscapes as the daytime labeled dataset in both supervised and unsupervised experiments.

\subsubsection{NightCity~\cite{tan2021night}} 
The NightCity is a large dataset of nighttime city driving scenes with  pixel-level annotations, which can be used for supervised semantic segmentation. There are 2,998 images for traning, 1,299 images for validation or testing with pixel-level annotations of 19 categories. The labeled object classes are the same as the Cityscapes~\cite{Cordts2016Cityscapes}.

\subsubsection{ACDC~\cite{sakaridis2021acdc}} 
ACDC is an adverse conditions dataset with the correspondences for semantic driving scene understanding relationship. It contains 4,006 images with high-quality pixel-level semantic annotations evenly distributed among the four common adverse conditions in real-world driving environments, namely fog, nighttime, rain and snow. Both the resolution and labeled categories are the same as Cityscapes~\cite{Cordts2016Cityscapes}. The ACDC dataset contains 1,000 haze images, 1,006 nighttime images, 1,000 rain and 1,000 snow images for dense pixel-level semantic annotation. We use the ACDC\_night as our supervised experimental dataset, which consists of 400 training, 106 validation and 500 test images.

\subsubsection{Dark Zurich~\cite{GCMA_UAE}} 
Dark Zurich is a large dataset with urban driving scenes designed for unsupervised semantic segmentation. It includes 2,416 nighttime images, 2,920 twilight images and 3,041 daytime images for training, which are all unlabeled with resolution of $1,920 \times 1,080$. 
These images are captured in the same scenes during daytime, twilight and nighttime so that they can be aligned by image features.
In this work, we only employ 2,416 night-day image pairs to train our unsupervised model. 
There are also 201 nighttime images with pixel-annotation in the Dark Zurich dataset, involving 50 images for validation (Dark Zurich-val) and 151 images for testing (Dark Zurich-test), which can be used for quantitative evaluation.
The Dark Zurich-test dataset provides only one validation channel via the official website.
We obtain the mIoU result of our proposed approach on Dark Zurich-test by submitting the segmentation predictions to the online evaluation website. 

\subsubsection{Nighttime Driving~\cite{dai2018dark}} 
The Nighttime Driving dataset ~\cite{dai2018dark} includes 50 nighttime driving-scene images with resolution of  $1,920 \times 1,080$. As in ~\cite{Cordts2016Cityscapes}, the images in  this set are all labeled with the same 19 classes. 
In this work, we adopt the Nighttime Driving dataset only for testing.

\subsection{Supervised Segmentation with DIAL-Filters}
\subsubsection{Experimental Setup}


We adopt several typical backbone networks, including DeepLabV2~\cite{chen2017deeplab}, RefineNet~\cite{lin2017refinenet} and PSPNet~\cite{zhao2017pspnet}, to verify the generalization capability of DIAL-Filters. Following~\cite{wu2021dannet}, all experiments utilize the semantic segmentation models that are pre-trained on Cityscapes for 150,000 epochs. The mIoU of pre-trained DeepLabV2, RefineNet and PSPNet on Cityscapes validation set are 66.37, 65.85 and 63.94, respectively. During training, we employ random cropping with size of $512\times512$ on the scale between 0.5 and 1.0, and  apply random horizontal flipping to expand the training dataset. As in~\cite{chen2017deeplab, wu2021dannet}, we train our model using the Stochastic Gradient Descent (SGD) optimizer with a momentum of 0.9 and a weight decay of $5 \times 10^{-4}$. 
The initial learning rate is set to $2.5 \times 10^{-4}$, and then we employ the poly learning rate policy to decrease it with a power of 0.9. The batch size is set to 4. We conducted our experiments on Tesla V100 GPU, and our approach is implemented by PyTorch.

\begin{table}
\centering
\caption{Comparison of our method and baseline models on the NightCity test set. ``C'': Trained on Cityscapes. ``C + N'': Trained on Cityscapes and NightCity. ``N\_t'': NightCity test set. ``C\_v'': Cityscapes validation set. \label{table:nightcity_res}}
\resizebox{1.0\linewidth}{!}{
\begin{tabular}{c|*{3}{c|}c}
\hline
\multirow{3}*{Methods} &\multicolumn{2}{c}{mIoU (\%) on N\_t} &\multicolumn{2}{|c}{mIoU (\%) on C\_v}\\
\cline{2-5}& & & &  \\[-6pt]
&{C}&{C + N}&{C}&{C + N}\\
\hline
  DeepLabV2~\cite{chen2017deeplab} & 18.20     & 46.39 &  66.37  & 65.65\\
  Ours (DeepLabV2) & -    & 48.24 &  - & 66.57\\
  \hline
  
  PSPNet~\cite{zhao2017pspnet} & 20.65    & 47.29 &  63.94  & 65.87\\
  Ours (PSPNet) & -     & 49.73 &  -  & 66.59\\
  \hline
  RefineNet~\cite{lin2017refinenet} & 22.92   & 48.70 &  65.85  & 65.68\\
  Ours (RefineNet)& -      &\textbf{ 51.21} &  -  & \textbf{67.15}\\
\hline

\hline
\end{tabular}}
\end{table}

\begin{figure*}[t]
\centering
\includegraphics[width=0.95\textwidth]{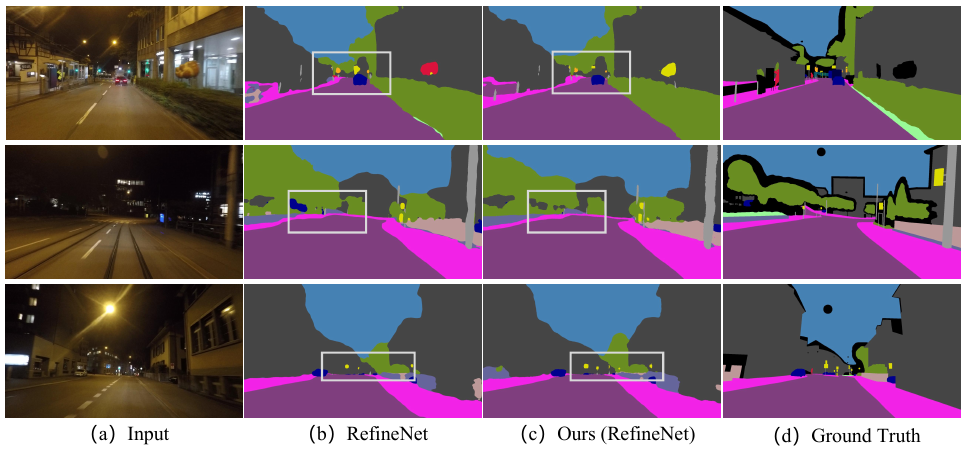} 
\caption{Visual segmentation results of our method and baseline model on ACDC\_night\_test images. All the methods are trainned on Cityscapes and ACDC\_night.}
\label{fig:acdc-visual}
\end{figure*}

\begin{figure}[t]
\centering
\includegraphics[width=0.5\textwidth]{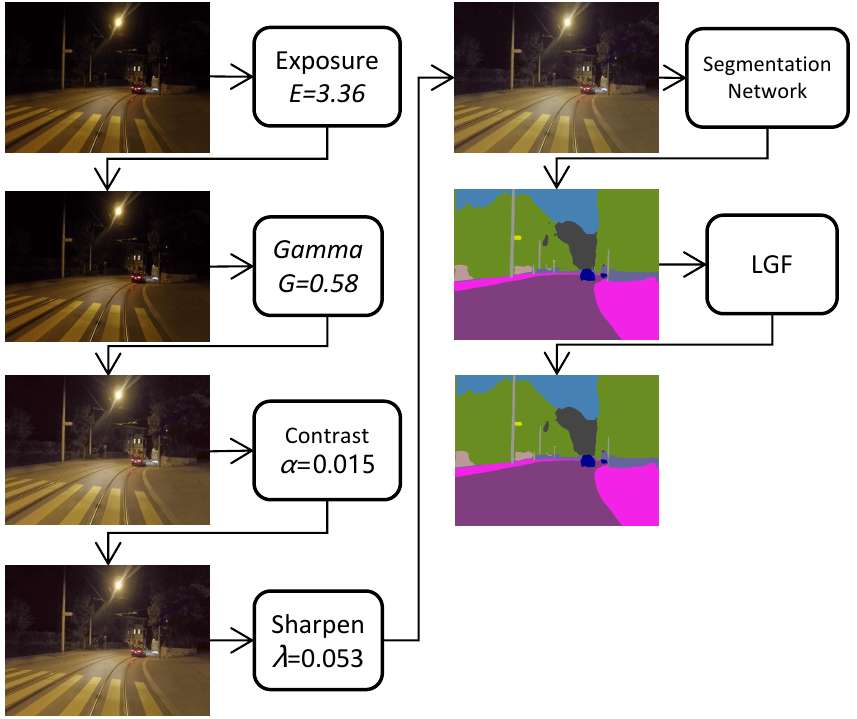} 
\caption{An example of the processing pipeline of our method. For better illustration, the filtered results are normalized. }
\label{fig:process-sequence}
\end{figure}

\subsubsection{Experiments on Cityscapes and NightCity Datasets}
To demonstrate the effectiveness of our proposed method, we plugin DIAL-Filters into three classic semantic segmentation networks, and perform experiments on three labeled datasets. Table~\ref{table:nightcity_res} reports the quantitative results of the existing methods and the proposed approach trained on Cityscapes (``C" columns) or hybrid datasets (``C+N" columns), respectively. With training on hybrid datasets (Cityscapes and NightCity) and validated on NightCity\_test, our method outperforms DeepLabV2, PSPNet and RefineNet by 1.85\%, 2.44\% and 2.41\%, respectively. Compared with these methods trained on daytime Cityscapes, our method can still improve them by 0.20\%, 2.65\% and 1.30\% on daytime Cityscapes validation set, while the baseline models of hybrid data training have less improvement or even become worse. This demonstrates that the IAPM module is able to adaptively process the image with different illumination for better semantic segmentation. Figure~\ref{fig:nightcity_viusal} shows several visual examples of our method and the baseline PSPNet (trained on ``C+N"). It can be observed that our method has better segmentation performance on the  categories that are overlooked by other methods at night, such as pole and traffic sign.
\subsubsection{Experiments on Cityscapes and ACDC\_night Datasets}
We examine the effectiveness of the proposed method on the hybrid datasets of Cityscapes and ACDC\_night. As depicted in Table~\ref{table:acdc-res}, our proposed DIAL-Filters with any of the three backbones performs better than the baseline models on ACDC\_night test dataset. Figure~\ref{fig:acdc-visual} shows the qualitative comparisons between our method and the baseline RefineNet. It can be observed that the presented IPAM module is able to reveal more objects by adaptively increasing the brightness and contrast of input image, which are essential to semantic segmentation in the region of small objects. Figure~\ref{fig:process-sequence} illustrates how the CNN-PP module predicts DIF's parameters, including the detailed parameter values and the images processed by each sub-filter. After the input image is processed by the learned DIF module, more image details are revealed, which are conducive to the subsequent segmentation task.
\begin{figure*}[t]
\centering
\includegraphics[width=0.95\textwidth]{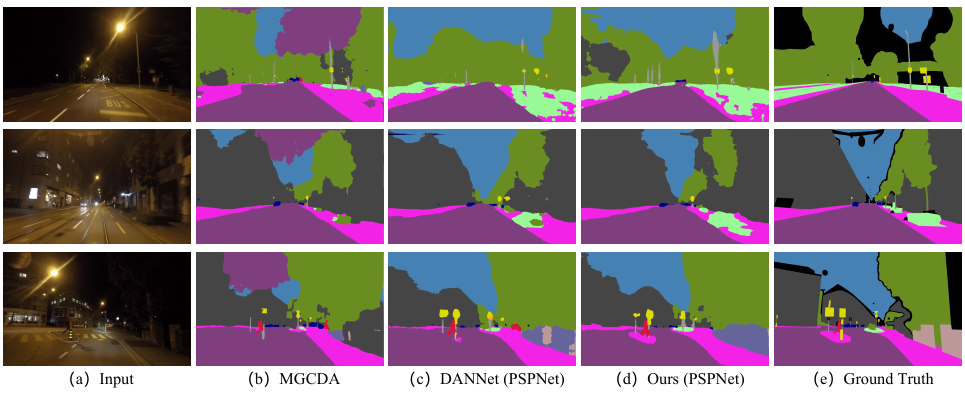} 
\caption{Qualitative comparisons of our approach with some methods on three samples from Dark Zurich-val. All the methods perform domain adaption from Cityscapes to Dark Zurich.}
\label{fig:DZ-visual}
\end{figure*}

\begin{table}
\centering
\caption{Comparison of our approach and baseline models on the ACDC\_night test set. ``C'': Trained on Cityscapes. ``C + A'': Trained on Cityscapes and ACDC\_night. ``A\_t'': ACDC\_night test set. ``C\_v'': Cityscapes validation set. \label{table:acdc-res}}
\resizebox{1.0\linewidth}{!}{
\begin{tabular}{c|*{3}{c|}c}
\hline
\multirow{3}*{Methods} &\multicolumn{2}{c}{mIoU (\%) on A\_t} &\multicolumn{2}{|c}{mIoU (\%) on C\_v}\\
\cline{2-5}& & & &  \\[-6pt]
&{C}&{C + A}&{C}&{C + A}\\
\hline
  DeepLabV2~\cite{chen2017deeplab} & 30.06     & 53.31 &  66.37  & 64.97\\
  Ours (DeepLabV2) & -    & 55.78 &  - & 65.55\\
  \hline
  PSPNet~\cite{zhao2017pspnet} & 26.62   & 56.69&  63.94  & 65.18\\
  Ours (PSPNet) & -     & 58.42 &  -  & \textbf{66.75}\\
  \hline
  RefineNet~\cite{lin2017refinenet} & 29.05  & 57.69 &  65.85  & 63.19\\
  Ours (RefineNet)& -      &\textbf{60.06} &  -  &66.19\\
\hline

\hline
\end{tabular}}
\end{table}

\subsection{Unsupervised Segmentation with DIAL-Filters}

\subsubsection{Experimental Setup}
As in the supervised experiments, we employ  DeepLabV2~\cite{chen2017deeplab}, RefineNet~\cite{lin2017refinenet} and PSPNet~\cite{zhao2017pspnet} as baseline models to perform unsupervised segmentation experiments. The proposed  model is trained by the Stochastic Gradient Descent (SGD) optimizer with a momentum of 0.9 and a weight decay of $5 \times 10^{-4}$. Like~\cite{wu2021dannet}, we employ Adam optimizer to train the discriminators with $\beta$ set to $(0.9,0.99)$. The learning rate of the discriminators is set to $2.5 \times 10^{-4}$. Moreover, we apply random cropping with the crop size of 512 on the scale between 0.5 and 1.0 for the Cityscapes dataset, and the crop size is set to 960 on the scale between 0.9 and 1.1 on the Dark Zurich dataset. In addition, random horizontal flipping is used in the training. The other related settings are consistent with the supervised experiments.

\begin{table*}[!ht]
	\centering
	\caption{The per-category results on Dark Zurich-test by current state-of-the-art methods and our method. Cityscapes$\rightarrow$DZ denotes the adaptation from  Cityscapes  to Dark Zurich-night. The best results are presented in {\bf BOLD}.}
	\vspace{4pt}
	\label{tab:DN-res}
	\renewcommand\arraystretch{1.1}
	\footnotesize
	\setlength{\tabcolsep}{0.66mm}{
		\begin{tabular}{lcccccccccccccccccccc}
			\hline
			Method  & \rotatebox{90}{road} & \rotatebox{90}{sidewalk} & \rotatebox{90}{building} & \rotatebox{90}{wall} & \rotatebox{90}{fence} & \rotatebox{90}{pole} & \rotatebox{90}{traffic light \ } & \rotatebox{90}{traffic sign} & \rotatebox{90}{vegetation} & \rotatebox{90}{terrain} & \rotatebox{90}{sky} & \rotatebox{90}{person} & \rotatebox{90}{rider} & \rotatebox{90}{car} & \rotatebox{90}{truck} & \rotatebox{90}{bus} & \rotatebox{90}{train} & \rotatebox{90}{motorcycle} & \rotatebox{90}{bicycle}  & \bf mIoU\\ 
			\hline
			RefineNet \cite{lin2017refinenet}-Cityscapes   &  68.8 & 23.2 & 46.8 & 20.8 & 12.6 & 29.8 & 30.4 & 26.9 & 43.1 & 14.3 & 0.3 & 36.9 & 49.7 & 63.6 & 6.8 & 0.2 & 24.0 & 33.6 & 9.3 & 28.5\\
			DeepLabV2  \cite{chen2017deeplab}-Cityscapes   &  79.0 & 21.8 & 53.0 & 13.3 & 11.2 & 22.5 & 20.2 & 22.1 & 43.5 & 10.4 & 18.0 & 37.4 & 33.8 & 64.1 & 6.4 & 0.0 & 52.3 & 30.4 & 7.4 & 28.8\\
			PSPNet \cite{zhao2017pspnet}-Cityscapes  & 78.2 & 19.0 &  51.2 &  15.5 &  10.6 &  30.3 & 28.9 &  22.0 &  56.7 &  13.3 &  20.8 &  38.2 &  21.8 &  52.1 &  1.6 &  0.0 & 53.2 &  23.2 &  10.7 & 28.8\\
			\hline
			AdaptSegNet-Cityscapes$\rightarrow$DZ \cite{tsai2018learning}  &  86.1 & 44.2 & 55.1 & 22.2 & 4.8 & 21.1 & 5.6 & 16.7 & 37.2 & 8.4 & 1.2 & 35.9 & 26.7 & 68.2 & 45.1 & 0.0 & 50.1 & 33.9 & 15.6 & 30.4 \\
			ADVENT-Cityscapes$\rightarrow$DZ \cite{vu2019advent} & 85.8 & 37.9 & 55.5 & 27.7 & 14.5 & 23.1 & 14.0 & 21.1 & 32.1 & 8.7 & 2.0 & 39.9 & 16.6 & 64.0 & 13.8 & 0.0 & 58.8 & 28.5 & 20.7 & 29.7 \\
			BDL-Cityscapes$\rightarrow$DZ \cite{li2019bidirectional}  & 85.3 & 41.1 & 61.9 & 32.7 & 17.4 & 20.6 & 11.4 & 21.3 & 29.4 & 8.9 & 1.1 & 37.4 & 22.1 & 63.2 & 28.2 & 0.0 & 47.7 & {\bf 39.4} & 15.7 & 30.8 \\
			DMAda \cite{dai2018dark}  & 75.5 & 29.1 & 48.6 & 21.3 & 14.3 & 34.3 & 36.8 & 29.9 & 49.4 & 13.8 & 0.4 & 43.3 & 50.2 & 69.4 & 18.4 & 0.0 & 27.6 & 34.9 & 11.9 & 32.1 \\
			GCMA \cite{sakaridis2019guided}  & 81.7 & 46.9 & 58.8 & 22.0 & 20.0 & 41.2 & {\bf 40.5} & {\bf 41.6} & 64.8 & 31.0 & 32.1 & {\bf 53.5} & 47.5 & {\bf 75.5} & 39.2 & 0.0 & 49.6 & 30.7 & 21.0 & 42.0 \\
			MGCDA \cite{sakaridis2020map} & 80.3 & 49.3 & 66.2 & 7.8 & 11.0 & {\bf 41.4} &  38.9 &  39.0 & 64.1 & 18.0 & 55.8 & \ 52.1 & {\bf 53.5} &  74.7 & {\bf 66.0} & 0.0 & 37.5 & 29.1 & 22.7 & 42.5 \\
			\hline
			DANNet (DeepLabV2)~\cite{wu2021dannet} & 88.6 & 53.4 & 69.8 & 34.0 & 20.0 & 25.0 & 31.5 & 35.9 &69.5 & \ 32.2 &  82.3 & 44.2 & 43.7 & 54.1 & 22.0 & 0.1 & 40.9 & 36.0 & {\bf24.1} & 42.5\\
			DANNet (RefineNet)~\cite{wu2021dannet} & 90.0 & 54.0 & {\bf 74.8} & {\bf 41.0} & 21.1 & 25.0 & 26.8 & 30.2 &  72.0 
			& 26.2 & {\bf 84.0} & 47.0 & 33.9 & 68.2 & 19.0 &  0.3 & 66.4 & 38.3 &  23.6 & 44.3\\
			DANNet (PSPNet)~\cite{wu2021dannet} &  90.4 &  60.1 & 71.0 & 33.6 & {\bf 22.9} & 30.6 & 34.3 & 33.7 & 70.5 & 31.8 & 80.2 & 45.7 & 41.6 & 67.4 & 16.8 & 0.0 & 73.0 & 31.6 & 22.9 &  45.2\\
			\hline
			Ours (DeepLabV2) & 88.7 & 55.8 & 69.8 & 34.7 & 17.1 & 31.7 & 26.6 & 34.4 &69.0 &  25.9 & 80.1 & 45.1 & 43.3 & 67.6 & 10.9 & {\bf 1.1} & 66.1 & 37.6 &  20.5& 43.5\\
			Ours (RefineNet) &  90.4 & {\bf 62.5} &  73.1 & 34.4 &  21.5 & 35.7 & 27.7 & 32.1 &  70.3 & {\bf 35.6} &  81.7 & 45.0 & 43.7 & 70.3 & 8.2 &  0.0 & 69.2 &38.0 & 18.2 & 45.1\\
			Ours (PSPNet) & {\bf 90.6} &  60.8 & 70.9 &  40.2 & 21.1 & 39.6 & 34.4 & 38.3 & {\bf 73.2} & 30.2 & 72.9 & 48.6 & 41.6 & 72.8 & 8.8 & 0.0 & {\bf 74.6} & 33.0 & 22.8 &  {\bf 46.0}\\
			\hline
	\end{tabular}}
\end{table*}

\begin{figure}[t]
\centering
\includegraphics[width=0.5\textwidth]{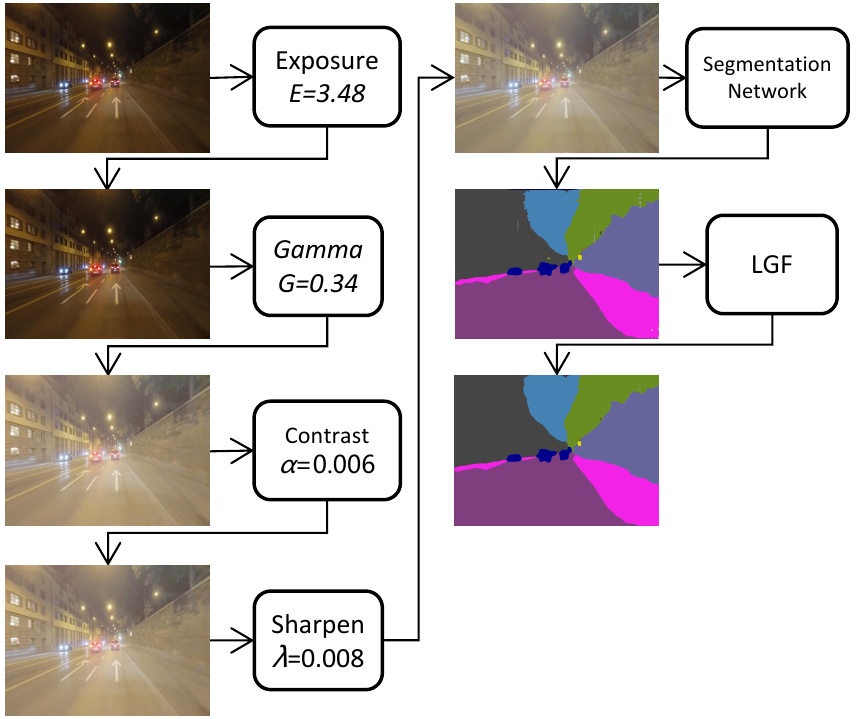} 
\caption{An example of the  processing pipeline of our method. For better illustration, the filtered results are normalized. }
\label{fig:process-sequence_un}
\end{figure}

\subsubsection{Comparison with state-of-the-art methods}
We compare the proposed approach with  state-of-the-art unsupervised segmentation methods, including DANNet~\cite{wu2021dannet}, MGCDA~\cite{sakaridis2020map}, GCMA~\cite{sakaridis2019guided}, DMAda~\cite{dai2018dark} and several domain adaptation methods~\cite{tsai2018learning,vu2019advent,li2019bidirectional} on Dark Zurich-test and Night Driving to demonstrate the efficacy of our method. All these competing methods adopt the  ResNet-101 backbone. Specifically, both our method and DANNet are tested with three baseline models. MGCDA, GCMA, and DMAda are tested with the baseline RefineNet, while the rest are based on DeepLabV2.

\emph{Experimental Results on Dark Zurich-test.} Table~\ref{tab:DN-res} reports the quantitative results on  Dark Zurich-test dataset. Comparing to the state-of-the-art nighttime segmentation methods, our proposed DIAL-Filters with PSPNet achieves the highest mIoU score. It is worthy mentioning that although our model is smaller, it outperforms DANNet on all the three baseline models. It can be found that our DIAL-Filters with either DeepLabV2, RefineNet or PSPNet achieves better performance than the domain adaptation methods (see the second panel in Table~\ref{tab:DN-res}). Figure~\ref{fig:DZ-visual} shows several visual comparison examples of MGCDA, DANNet and our method. With the proposed DIAL-Filters, our adaptive module is able to  distinguish objects of interest from the images, especially small objects and  confusing areas with mixed categories in the dark.
Figure~\ref{fig:process-sequence_un} shows an example on how the CNN-PP module predicts DIF's parameters, including the detailed parameter values and the images processed by each sub-filter. It can be observed that our proposed DIAL-Filters are able to increase the brightness of the input image and reveal the image details, which is essential to segment the nighttime images.


\emph{Experimental Results on Night Driving.} Table~\ref{tab:ND-test} reports the mIoU results on Night Driving test dataset. In contrast to the state-of-the-art nighttime segmentation methods, our DIAL-Filters with PSPNet achieves the best performance. Though our model is smaller, it outperforms DANNet by 2.21\%, 1.96\% and 2.62\%, respectively, when RefineNet, DeepLabV2 and PSPNet are used as baselines. In addition, it can be clearly seen that our method achieves better performance than the domain adaptation methods.

\begin{table}[!ht]
	\centering\footnotesize
	\caption{Comparison of our approach with the existing state-of-the-art methods on Nighttime Driving test set \cite{dai2018dark}.}
	\vspace{4pt}
	\label{tab:ND-test}
	\renewcommand\arraystretch{1.08}
	\setlength{\tabcolsep}{2mm}{
		\footnotesize
		\begin{tabular}{lc}
			\hline
			Method  &  mIoU\\ 
			\hline
			RefineNet \cite{lin2017refinenet}-Cityscapes  &32.75\\
			DeepLabV2  \cite{chen2017deeplab}-Cityscapes  &25.44\\
			PSPNet \cite{zhao2017pspnet}-Cityscapes  & 27.65\\
			\hline
			AdaptSegNet-Cityscapes$\rightarrow$DZ-night \cite{tsai2018learning}  &34.5 \\
			ADVENT-Cityscapes$\rightarrow$DZ-night \cite{vu2019advent}  &34.7\\
			BDL-Cityscapes$\rightarrow$DZ-night \cite{li2019bidirectional}  &34.7 \\
			DMAda \cite{dai2018dark}  &36.1 \\
			GCMA \cite{sakaridis2019guided}  &45.6 \\
			MGCDA \cite{sakaridis2020map} & 49.4 \\
			\hline
			DANNet (RefineNet) & 42.36\\
			DANNet (DeepLabV2) & 44.98\\
			DANNet (PSPNet) & 47.70 \\
			\hline
			Ours (RefineNet) & 44.57\\
			Ours (DeepLabV2) & 46.94\\
			Ours (PSPNet) &{\bf 50.32} \\
			\hline
	\end{tabular}}
	\vspace{-5pt}
\end{table}

\subsection{Ablation Study}

To examine the effectiveness of each module in our proposed framework, including IAPM, LGF and DIF, we conduct ablation experiments with different settings. All experiments are trained on the mixed datasets of Cityscapes and NightCity in a supervised manner, where the weight parameters are pre-trained 150,000 epochs on Cityscapes. 

\begin{figure}[t]
\centering
\includegraphics[width=0.48\textwidth]{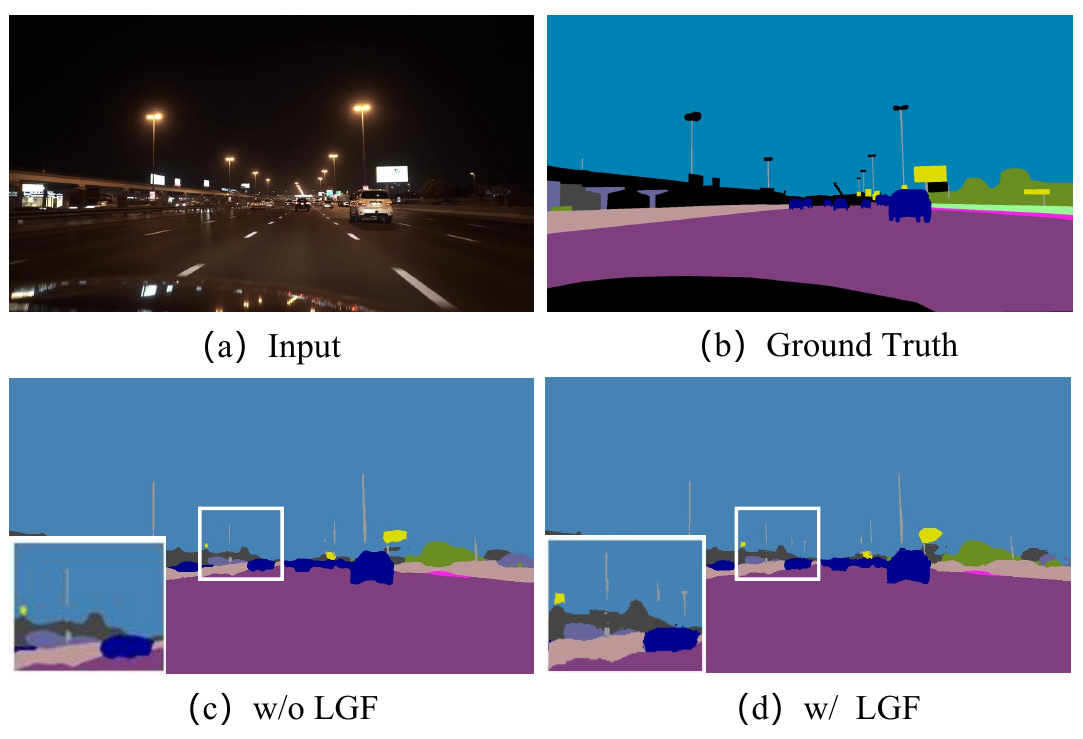}  \caption{Visual results of w/ and w/o the LGF module on a sample from NightCity\_test by our method (RefineNet).}
\label{fig:ablation}
\end{figure}

Table~\ref{tab:ablation} shows the experimental results. We select RefineNet (ResNet-101) as the base model, and 'DIAL-Filters' is the full model of our method. The settings and training data are the same for all experiments. It can be seen that DIF preprocessing, LGF postprocessing and image adaptive IAPM all improve the segmentation performance. The RefineNet\_deep is a deeper version of RefineNet, whose backbone is ResNet-152 with 15,644K more learnable parameters than ResNet-101. Our proposed approach performs better than RefineNet\_deep with only 280K additional parameters in CNN-PP and LGF. The method with fixed DIF means that the filter's hyperparameters are a set of  given values, all of which are within a reasonable range. Clearly, our DIAL-Filters method achieves the best performance on both NightCity\_test and Cityscapes\_test, which indicates that our method can adaptively process both the daytime images and nighttime ones. This is essential to the down-streamed segmentation tasks. Moreover, the LGF for postprocessing can further boost the performance. Figure~\ref{fig:ablation} shows the visual results with/without LGF. It can be seen that the learnable guided filter obtains more precise segmentation boundaries of small objects. We also evaluate the selection of the proposed differentiable filters on the testing datasets. As shown in Table~\ref{tab:ablation}, without any one of our proposed four filters, the performance is deteriorated while still outperforming both the fixed DIF and the original baseline. This further demonstrates the effectiveness of our proposed differentiable filters and adaptive processing strategy. 

\begin{table}[ht]
	\centering
	\caption{Ablation study on the variants of our DIAL-Filters (RefineNet) on NightCity\_test and Cityscaes\_val.} 
	\vspace{4pt}
	\label{tab:ablation}
	\renewcommand\arraystretch{1.08}
	\setlength{\tabcolsep}{2mm}{
		\small\footnotesize
		\begin{tabular}{lcc}
			\hline
			Method  &  mIoU (\%) on N\_t & mIoU (\%) on C\_v \\ 
			\hline
			RefineNet &48.70&65.68\\
			\hline
			RefineNet\_deep  & 49.80&66.43\\
			\hline
            w/o Exposure Filter & 50.28&66.00\\
       	w/o Gamma Filter & 49.69&66.10\\
       	w/o Contrast Filter & 50.36 &65.95\\
   		w/o Sharpen Filter & 50.43 &66.09\\
			\hline
   		w/o IAPM & 48.86&66.33\\
			\hline
			w/o LGF  &  50.97&65.93\\
			\hline
			w/ fixed DIF  &49.17&65.77 \\
			\hline
			DIAL-Filters & {\bf 51.21}&{\bf 67.15}\\
			\hline
	\end{tabular}}
	\vspace{-5pt}
\end{table}

\begin{table}[!ht]
\centering
\caption{Efficiency analysis on the compared methods.}
\label{tab:Efficiency}
\begin{tabular}{ccc}
    \hline
  
    Method&Additional Params&Speed (ms)\\
    \cline{1-3}
    RefineNet &/&20\\
    RefineNet\_deep &15,644K&27\\
    DANNet (RefineNet)&4,299K&23\\
    Ours (RefineNet)&280K&24\\

  \hline
\end{tabular}
\end{table}

\subsection{Efficiency Analysis}
In our proposed framework, we introduce a set of novel learnable DIAL-Filters with 280K trainable parameters into a segmentation network. CNN-PP has five convolutional layers, a dropout layer and a fully connected layer, and LGF includes two convolutional layers. Based on RefineNet, Table \ref{tab:Efficiency} compares the efficiency of some methods used in our experiments. All these methods deploy an add-on module into RefineNet. The second column lists the number of additional parameters over the RefineNet model. The third column lists the running time on a color image with of size $512 \times 1024$ using a single Tesla V100 GPU. It can be observed that our method only adds 280K trainable parameters over RefineNet while achieving the best performance on all experiments with comparable running time. Note that though our method has fewer trainable parameters than DANNet, its running time is slightly longer. This is because the filtering process in the DIF module incurs extra computation.

\section{Conclusion} 
In this paper, we proposed a novel approach to semantic segmentation in nighttime driving conditions by adaptively enhancing each input image to obtain better performance. Specifically, we introduced dual image-adaptive learnable filters (DIAL-Filters) and embedded them into the head and end of a segmentation network. A fully differentiable image processing module was developed to preprocess the input image, whose hyperparameters were predicted by a small convolutional neural network. 
The preliminary segmentation results were further enhanced by learnable guided filtering for more accurate segmentation.
The whole framework was trained in an end-to-end fashion, where the parameter prediction network was weakly supervised to learn an appropriate DIF module through the segmentation loss in the supervised experiments. 
Our experiments on both supervised  and unsupervised segmentation demonstrated the superiority of the proposed DIAL-Filters to previous nighttime driving-scene semantic segmentation methods.


\ifCLASSOPTIONcaptionsoff
  \newpage
\fi

\bibliographystyle{IEEEtran}
\bibliography{ieeetran}

\begin{thebibliography}{10}
\providecommand{\url}[1]{#1}
\csname url@samestyle\endcsname
\providecommand{\newblock}{\relax}
\providecommand{\bibinfo}[2]{#2}
\providecommand{\BIBentrySTDinterwordspacing}{\spaceskip=0pt\relax}
\providecommand{\BIBentryALTinterwordstretchfactor}{4}
\providecommand{\BIBentryALTinterwordspacing}{\spaceskip=\fontdimen2\font plus
\BIBentryALTinterwordstretchfactor\fontdimen3\font minus
  \fontdimen4\font\relax}
\providecommand{\BIBforeignlanguage}[2]{{%
\expandafter\ifx\csname l@#1\endcsname\relax
\typeout{** WARNING: IEEEtran.bst: No hyphenation pattern has been}%
\typeout{** loaded for the language `#1'. Using the pattern for}%
\typeout{** the default language instead.}%
\else
\language=\csname l@#1\endcsname
\fi
#2}}
\providecommand{\BIBdecl}{\relax}
\BIBdecl

\bibitem{yang2018real}
L.~Yang, X.~Liang, T.~Wang, and E.~Xing, ``Real-to-virtual domain unification
  for end-to-end autonomous driving,'' in \emph{Proceedings of the European
  Conference on Computer Vision}, 2018, pp. 530--545.

\bibitem{teso2020semantic}
D.~Teso-Fz-Beto{\~n}o, E.~Zulueta, A.~S{\'a}nchez-Chica, U.~Fernandez-Gamiz,
  and A.~Saenz-Aguirre, ``Semantic segmentation to develop an indoor navigation
  system for an autonomous mobile robot,'' \emph{Mathematics}, vol.~8, no.~5,
  p. 855, 2020.

\bibitem{thomas2021self}
H.~Thomas, B.~Agro, M.~Gridseth, J.~Zhang, and T.~D. Barfoot, ``Self-supervised
  learning of lidar segmentation for autonomous indoor navigation,'' in
  \emph{2021 IEEE International Conference on Robotics and Automation
  (ICRA)}.\hskip 1em plus 0.5em minus 0.4em\relax IEEE, 2021, pp.
  14\,047--14\,053.

\bibitem{Li2016Combining}
C.~Li and M.~Wand, ``Combining markov random fields and convolutional neural
  networks for image synthesis,'' in \emph{Proceedings of the IEEE/CVF
  Conference on Computer Vision and Pattern Recognition}, 2016, pp. 2479--2486.

\bibitem{he2016deep}
K.~He, X.~Zhang, S.~Ren, and J.~Sun, ``Deep residual learning for image
  recognition,'' in \emph{Proceedings of the IEEE/CVF Conference on Computer
  Vision and Pattern Recognition}, 2016, pp. 770--778.

\bibitem{lin2017refinenet}
G.~Lin, A.~Milan, C.~Shen, and I.~Reid, ``Refinenet: Multi-path refinement
  networks for high-resolution semantic segmentation,'' in \emph{Proceedings of
  the IEEE/CVF Conference on Computer Vision and Pattern Recognition}, 2017,
  pp. 1925--1934.

\bibitem{zhao2017pspnet}
H.~Zhao, J.~Shi, X.~Qi, X.~Wang, and J.~Jia, ``Pyramid scene parsing network,''
  in \emph{Proceedings of the IEEE/CVF Conference on Computer Vision and
  Pattern Recognition}, 2017, pp. 2881--2890.

\bibitem{weng2021stage}
X.~Weng, Y.~Yan, S.~Chen, J.-H. Xue, and H.~Wang, ``Stage-aware feature
  alignment network for real-time semantic segmentation of street scenes,''
  \emph{IEEE Transactions on Circuits and Systems for Video Technology},
  vol.~32, no.~7, pp. 4444--4459, 2021.

\bibitem{ji2020encoder}
J.~Ji, R.~Shi, S.~Li, P.~Chen, and Q.~Miao, ``Encoder-decoder with cascaded
  crfs for semantic segmentation,'' \emph{IEEE Transactions on Circuits and
  Systems for Video Technology}, vol.~31, no.~5, pp. 1926--1938, 2020.

\bibitem{sun2021gaussian}
X.~Sun, C.~Chen, X.~Wang, J.~Dong, H.~Zhou, and S.~Chen, ``Gaussian dynamic
  convolution for efficient single-image segmentation,'' \emph{IEEE
  Transactions on Circuits and Systems for Video Technology}, vol.~32, no.~5,
  pp. 2937--2948, 2021.

\bibitem{geiger2012we}
A.~Geiger, P.~Lenz, and R.~Urtasun, ``Are we ready for autonomous driving? the
  kitti vision benchmark suite,'' in \emph{Proceedings of the IEEE/CVF
  Conference on Computer Vision and Pattern Recognition}, 2012, pp. 3354--3361.

\bibitem{Cordts2016Cityscapes}
M.~Cordts, M.~Omran, S.~Ramos, T.~Rehfeld, M.~Enzweiler, R.~Benenson,
  U.~Franke, S.~Roth, and B.~Schiele, ``The cityscapes dataset for semantic
  urban scene understanding,'' in \emph{Proceedings of the IEEE Conference on
  Computer Vision and Pattern Recognition}, 2016, pp. 3213--3223.

\bibitem{tan2021night}
X.~Tan, K.~Xu, Y.~Cao, Y.~Zhang, L.~Ma, and R.~W. Lau, ``Night-time scene
  parsing with a large real dataset,'' \emph{IEEE Transactions on Image
  Processing}, vol.~30, pp. 9085--9098, 2021.

\bibitem{sakaridis2021acdc}
C.~Sakaridis, D.~Dai, and L.~Van~Gool, ``Acdc: The adverse conditions dataset
  with correspondences for semantic driving scene understanding,'' in
  \emph{Proceedings of the IEEE/CVF International Conference on Computer
  Vision}, 2021, pp. 10\,765--10\,775.

\bibitem{wu2021dannet}
X.~Wu, Z.~Wu, H.~Guo, L.~Ju, and S.~Wang, ``Dannet: A one-stage domain
  adaptation network for unsupervised nighttime semantic segmentation,'' in
  \emph{Proceedings of the IEEE/CVF Conference on Computer Vision and Pattern
  Recognition}, 2021, pp. 15\,769--15\,778.

\bibitem{dai2018dark}
D.~Dai and L.~Van~Gool, ``Dark model adaptation: Semantic image segmentation
  from daytime to nighttime,'' in \emph{IEEE Intelligent Transportation Systems
  Conference}, 2018, pp. 3819--3824.

\bibitem{sakaridis2019guided}
C.~Sakaridis, D.~Dai, and L.~V. Gool, ``Guided curriculum model adaptation and
  uncertainty-aware evaluation for semantic nighttime image segmentation,'' in
  \emph{Proceedings of the IEEE/CVF International Conference on Computer
  Vision}, 2019, pp. 7374--7383.

\bibitem{sun2019see}
L.~Sun, K.~Wang, K.~Yang, and K.~Xiang, ``See clearer at night: towards robust
  nighttime semantic segmentation through day-night image conversion,'' in
  \emph{Artificial Intelligence and Machine Learning in Defense Applications},
  vol. 11169.\hskip 1em plus 0.5em minus 0.4em\relax International Society for
  Optics and Photonics, 2019, p. 111690A.

\bibitem{nag2019s}
S.~Nag, S.~Adak, and S.~Das, ``What’s there in the dark,'' in
  \emph{International Conference on Image Processing}.\hskip 1em plus 0.5em
  minus 0.4em\relax IEEE, 2019, pp. 2996--3000.

\bibitem{sakaridis2020map}
C.~Sakaridis, D.~Dai, and L.~Van~Gool, ``Map-guided curriculum domain
  adaptation and uncertainty-aware evaluation for semantic nighttime image
  segmentation,'' \emph{IEEE Transactions on Pattern Analysis and Machine
  Intelligence}, vol.~44, no.~6, pp. 3139--3153, 2020.

\bibitem{hu2018exposure}
Y.~Hu, H.~He, C.~Xu, B.~Wang, and S.~Lin, ``Exposure: A white-box photo
  post-processing framework,'' \emph{ACM Transactions on Graphics (TOG)},
  vol.~37, no.~2, p.~26, 2018.

\bibitem{yu2018deepexposure}
R.~Yu, W.~Liu, Y.~Zhang, Z.~Qu, D.~Zhao, and B.~Zhang, ``Deepexposure: Learning
  to expose photos with asynchronously reinforced adversarial learning,'' in
  \emph{Proceedings of the 32nd International Conference on Neural Information
  Processing Systems}, 2018, pp. 2153--2163.

\bibitem{zeng2020learning}
H.~Zeng, J.~Cai, L.~Li, Z.~Cao, and L.~Zhang, ``Learning image-adaptive 3d
  lookup tables for high performance photo enhancement in real-time,''
  \emph{IEEE Transactions on Pattern Analysis and Machine Intelligence},
  vol.~44, no.~4, pp. 2058--2073, 2020.

\bibitem{liu2022image}
W.~Liu, G.~Ren, R.~Yu, S.~Guo, J.~Zhu, and L.~Zhang, ``Image-adaptive yolo for
  object detection in adverse weather conditions,'' in \emph{Proceedings of the
  AAAI Conference on Artificial Intelligence}, vol.~36, no.~2, 2022, pp.
  1792--1800.

\bibitem{Long2015Fully}
J.~Long, E.~Shelhamer, and T.~Darrell, ``Fully convolutional networks for
  semantic segmentation,'' in \emph{Proceedings of the IEEE/CVF Conference on
  Computer Vision and Pattern Recognition}, 2015, pp. 3431--3440.

\bibitem{chen2017deeplab}
L.-C. Chen, G.~Papandreou, I.~Kokkinos, K.~Murphy, and A.~L. Yuille, ``Deeplab:
  Semantic image segmentation with deep convolutional nets, atrous convolution,
  and fully connected crfs,'' \emph{PAMI}, vol.~40, no.~4, pp. 834--848, 2017.

\bibitem{chen2017rethinking}
L.-C. Chen, G.~Papandreou, F.~Schroff, and H.~Adam, ``Rethinking atrous
  convolution for semantic image segmentation,'' \emph{arXiv preprint
  arXiv:1706.05587}, 2017.

\bibitem{chen2018encoder}
L.-C. Chen, Y.~Zhu, G.~Papandreou, F.~Schroff, and H.~Adam, ``Encoder-decoder
  with atrous separable convolution for semantic image segmentation,'' in
  \emph{Proceedings of the European Conference on Computer Vision}, 2018, pp.
  801--818.

\bibitem{polesel2000image}
A.~Polesel, G.~Ramponi, and V.~J. Mathews, ``Image enhancement via adaptive
  unsharp masking,'' \emph{IEEE Transactions on Image Processing}, vol.~9,
  no.~3, pp. 505--510, 2000.

\bibitem{yu2004fast}
Z.~Yu and C.~Bajaj, ``A fast and adaptive method for image contrast
  enhancement,'' in \emph{2004 International Conference on Image Processing,
  2004. ICIP'04.}, vol.~2.\hskip 1em plus 0.5em minus 0.4em\relax IEEE, 2004,
  pp. 1001--1004.

\bibitem{wang2021adaptive}
W.~Wang, Z.~Chen, X.~Yuan, and F.~Guan, ``An adaptive weak light image
  enhancement method,'' in \emph{Twelfth International Conference on Signal
  Processing Systems}, vol. 11719.\hskip 1em plus 0.5em minus 0.4em\relax
  International Society for Optics and Photonics, 2021, p. 1171902.

\bibitem{zhang2015image}
L.~Zhang, ``Image adaptive edge detection based on canny operator and
  multiwavelet denoising,'' in \emph{2014 International Conference on Computer
  Science and Electronic Technology}.\hskip 1em plus 0.5em minus 0.4em\relax
  Atlantis Press, 2015, pp. 335--338.

\bibitem{tian2021partial}
Y.~Tian and S.~Zhu, ``Partial domain adaptation on semantic segmentation,''
  \emph{IEEE Transactions on Circuits and Systems for Video Technology},
  vol.~32, no.~6, pp. 3798--3809, 2021.

\bibitem{zhang2018unsupervised}
L.~Zhang, P.~Wang, W.~Wei, H.~Lu, C.~Shen, A.~van~den Hengel, and Y.~Zhang,
  ``Unsupervised domain adaptation using robust class-wise matching,''
  \emph{IEEE Transactions on Circuits and Systems for Video Technology},
  vol.~29, no.~5, pp. 1339--1349, 2018.

\bibitem{lin2020cross}
K.~Lin, L.~Wang, K.~Luo, Y.~Chen, Z.~Liu, and M.-T. Sun, ``Cross-domain
  complementary learning using pose for multi-person part segmentation,''
  \emph{IEEE Transactions on Circuits and Systems for Video Technology},
  vol.~31, no.~3, pp. 1066--1078, 2020.

\bibitem{bo2022all}
Q.~Bo, W.~Ma, Y.-K. Lai, and H.~Zha, ``All-higher-stages-in adaptive context
  aggregation for semantic edge detection,'' \emph{IEEE Transactions on
  Circuits and Systems for Video Technology}, vol.~32, no.~10, pp. 6778--6791,
  2022.

\bibitem{tsai2018learning}
Y.-H. Tsai, W.-C. Hung, S.~Schulter, K.~Sohn, M.-H. Yang, and M.~Chandraker,
  ``Learning to adapt structured output space for semantic segmentation,'' in
  \emph{Proceedings of the IEEE/CVF Conference on Computer Vision and Pattern
  Recognition}, 2018, pp. 7472--7481.

\bibitem{pan2020unsupervised}
F.~Pan, I.~Shin, F.~Rameau, S.~Lee, and I.~S. Kweon, ``Unsupervised
  intra-domain adaptation for semantic segmentation through self-supervision,''
  in \emph{Proceedings of the IEEE/CVF Conference on Computer Vision and
  Pattern Recognition}, 2020, pp. 3764--3773.

\bibitem{vu2019advent}
T.-H. Vu, H.~Jain, M.~Bucher, M.~Cord, and P.~P{\'e}rez, ``Advent: Adversarial
  entropy minimization for domain adaptation in semantic segmentation,'' in
  \emph{Proceedings of the IEEE/CVF Conference on Computer Vision and Pattern
  Recognition}, 2019, pp. 2517--2526.

\bibitem{xie2022sepico}
B.~Xie, S.~Li, M.~Li, C.~H. Liu, G.~Huang, and G.~Wang, ``Sepico:
  Semantic-guided pixel contrast for domain adaptive semantic segmentation,''
  \emph{arXiv preprint arXiv:2204.08808}, 2022.

\bibitem{zhang2021prototypical}
P.~Zhang, B.~Zhang, T.~Zhang, D.~Chen, Y.~Wang, and F.~Wen, ``Prototypical
  pseudo label denoising and target structure learning for domain adaptive
  semantic segmentation,'' in \emph{Proceedings of the IEEE/CVF Conference on
  Computer Vision and Pattern Recognition}, 2021, pp. 12\,414--12\,424.

\bibitem{li2019bidirectional}
Y.~Li, L.~Yuan, and N.~Vasconcelos, ``Bidirectional learning for domain
  adaptation of semantic segmentation,'' in \emph{Proceedings of the IEEE/CVF
  Conference on Computer Vision and Pattern Recognition}, 2019, pp. 6936--6945.

\bibitem{zhang2017curriculum}
Y.~Zhang, P.~David, and B.~Gong, ``Curriculum domain adaptation for semantic
  segmentation of urban scenes,'' in \emph{Proceedings of the IEEE
  International Conference on Computer Vision}, 2017, pp. 2020--2030.

\bibitem{lian2019constructing}
Q.~Lian, F.~Lv, L.~Duan, and B.~Gong, ``Constructing self-motivated pyramid
  curriculums for cross-domain semantic segmentation: A non-adversarial
  approach,'' in \emph{Proceedings of the IEEE/CVF International Conference on
  Computer Vision}, 2019, pp. 6758--6767.

\bibitem{richter2016playing}
S.~R. Richter, V.~Vineet, S.~Roth, and V.~Koltun, ``Playing for data: Ground
  truth from computer games,'' in \emph{European Conference on Computer
  Vision}.\hskip 1em plus 0.5em minus 0.4em\relax Springer, 2016, pp. 102--118.

\bibitem{chen2017no}
Y.-H. Chen, W.-Y. Chen, Y.-T. Chen, B.-C. Tsai, Y.-C. Frank~Wang, and M.~Sun,
  ``No more discrimination: Cross city adaptation of road scene segmenters,''
  in \emph{Proceedings of the IEEE International Conference on Computer
  Vision}, 2017, pp. 1992--2001.

\bibitem{romera2019bridging}
E.~Romera, L.~M. Bergasa, K.~Yang, J.~M. Alvarez, and R.~Barea, ``Bridging the
  day and night domain gap for semantic segmentation,'' in \emph{2019 IEEE
  Intelligent Vehicles Symposium}, 2019, pp. 1312--1318.

\bibitem{zhu2017unpaired}
J.~Y. Zhu, T.~Park, P.~Isola, and A.~A. Efros, ``Unpaired image-to-image
  translation using cycle-consistent adversarial networks,'' in
  \emph{Proceedings of the IEEE/CVF International Conference on Computer
  Vision}, 2017, pp. 2242--2251.

\bibitem{wu2021one}
X.~Wu, Z.~Wu, L.~Ju, and S.~Wang, ``A one-stage domain adaptation network with
  image alignment for unsupervised nighttime semantic segmentation,''
  \emph{IEEE Transactions on Pattern Analysis and Machine Intelligence},
  no.~01, pp. 1--1, 2021.

\bibitem{mosleh2020hardware}
A.~Mosleh, A.~Sharma, E.~Onzon, F.~Mannan, N.~Robidoux, and F.~Heide,
  ``Hardware-in-the-loop end-to-end optimization of camera image processing
  pipelines,'' in \emph{Proceedings of the IEEE/CVF Conference on Computer
  Vision and Pattern Recognition}, 2020, pp. 7529--7538.

\bibitem{hu2017deep}
P.~Hu, B.~Shuai, J.~Liu, and G.~Wang, ``Deep level sets for salient object
  detection,'' in \emph{Proceedings of the IEEE Conference on Computer Vision
  and Pattern Recognition}, 2017, pp. 2300--2309.

\bibitem{wu2018fast}
H.~Wu, S.~Zheng, J.~Zhang, and K.~Huang, ``Fast end-to-end trainable guided
  filter,'' in \emph{Proceedings of the IEEE Conference on Computer Vision and
  Pattern Recognition}, 2018, pp. 1838--1847.

\bibitem{he2012guided}
K.~He, J.~Sun, and X.~Tang, ``Guided image filtering,'' \emph{IEEE transactions
  on pattern analysis and machine intelligence}, vol.~35, no.~6, pp.
  1397--1409, 2012.

\bibitem{GCMA_UAE}
C.~Sakaridis, D.~Dai, , and L.~{Van Gool}, ``Guided curriculum model adaptation
  and uncertainty-aware evaluation for semantic nighttime image segmentation,''
  in \emph{Proceedings of the IEEE/CVF International Conference on Computer
  Vision}, 2019.

\bibitem{mao2017least}
X.~Mao, Q.~Li, H.~Xie, R.~Y. Lau, Z.~Wang, and S.~Paul~Smolley, ``Least squares
  generative adversarial networks,'' in \emph{Proceedings of the IEEE/CVF
  Conference on Computer Vision and Pattern Recognition}, 2017, pp. 2794--2802.

\end{thebibliography}

\end{document}